\documentclass[10pt,twocolumn,letterpaper]{article}

\usepackage{iccv}
\usepackage{times}
\usepackage{epsfig}
\usepackage{graphicx}
\usepackage{amsmath}
\usepackage{amssymb}

\usepackage{paralist}
\usepackage{bm}
\usepackage{amsfonts}
\usepackage{multirow}
\usepackage{booktabs}
\usepackage{colortbl}
\usepackage{setspace}

\usepackage[pagebackref=true,breaklinks=true,colorlinks=true,bookmarks=false]{hyperref} 

\usepackage[accsupp]{axessibility}  

\iccvfinalcopy 

\def\iccvPaperID{****} 

\begin{document}

\title{Grounded Entity-Landmark Adaptive Pre-training \\
for Vision-and-Language Navigation}

\author{Yibo Cui$^{1,2}$,~~\hspace{1pt}Liang Xie$^{*1,2}$,~~\hspace{1pt}Yakun Zhang$^{1,2}$,~~\hspace{1pt}Meishan Zhang$^{3}$,~~\hspace{1pt}Ye Yan$^{1,2}$,~~\hspace{1pt}Erwei Yin$^{*1,2}$ \\
	{$^1$} Defense Innovation Institute, Chinese Academy of Military Science \\
	{$^2$}  Tianjin Artificial Intelligence Innovation Center
	~~\hspace{1pt}{$^3$}  Harbin Institute of Technology (Shenzhen) \\
	\small \url{https://github.com/CSir1996/VLN-GELA}
	%
}

\maketitle

\begin{abstract}
Cross-modal alignment is one key challenge for Vision-and-Language Navigation (VLN). 
Most existing studies concentrate on mapping the global instruction or single sub-instruction to the corresponding trajectory. 
However, another critical problem of achieving fine-grained alignment at the entity level is seldom considered. 
To address this problem, we propose a novel Grounded Entity-Landmark Adaptive (GELA)  pre-training paradigm for VLN tasks. 
To achieve the adaptive pre-training paradigm, we first introduce grounded entity-landmark human annotations into the Room-to-Room (R2R) dataset, named GEL-R2R. 
Additionally, we adopt three grounded entity-landmark adaptive pre-training objectives: 1) entity phrase prediction, 2) landmark bounding box prediction, and 3) entity-landmark semantic alignment, which explicitly supervise the learning of fine-grained cross-modal alignment between entity phrases and environment landmarks.  
Finally, we validate our model on two downstream benchmarks: VLN with descriptive instructions (R2R) and dialogue instructions (CVDN). 
The comprehensive experiments show that our GELA model achieves state-of-the-art results on both tasks, demonstrating its effectiveness and generalizability.
\end{abstract}

{\let\thefootnote\relax\footnote{$^{*}$Corresponding author.}}

\section{Introduction}
\label{sec:intro}

\begin{figure}
    \centering
    \includegraphics[width=0.45\textwidth]{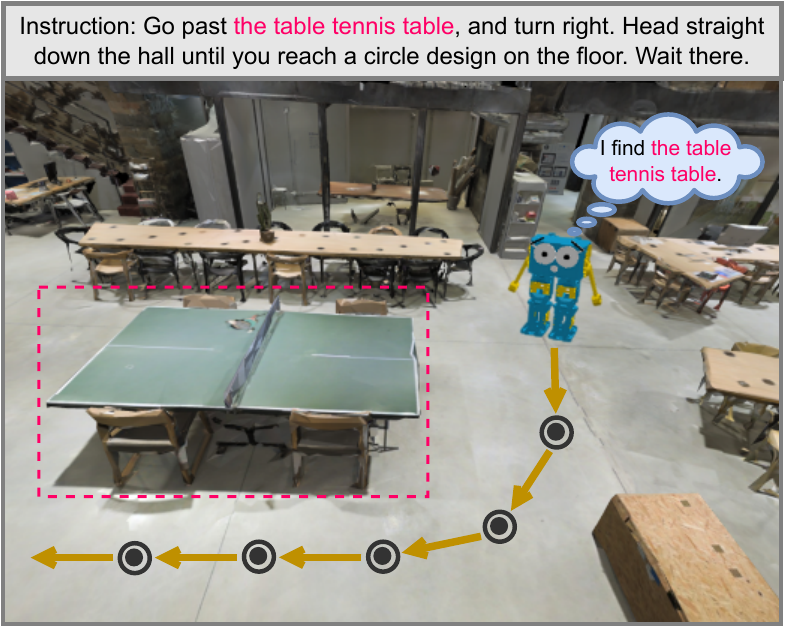}
    \caption{An illustration of an embodied agent navigating in a 3D photo-realistic environment. The agent is expected to navigate based on environment landmarks that correspond to the entity phrase in the instruction.}
    \label{fig:example}
\end{figure}

Vision-and-Language Navigation (VLN) \cite{DBLP:conf/cvpr/AndersonWTB0S0G18} is an important task in the Embodied Vision community, which has gained great attention \cite{DBLP:conf/cvpr/Zhu0CL20, DBLP:conf/eccv/MajumdarSLAPB20, DBLP:journals/corr/abs-2203-11591}.
It aims to ask an agent to reach the target location inside photo-realistic environments by following natural language instructions.
In VLN, cross-modal alignment is one critical step to accurately make the action decision for the agent \cite{DBLP:conf/cvpr/WangHcGSWWZ19}, since matching the mentioned landmarks (objects or scenes) with visual observations can help the comprehensive understanding of the environments and instructions \cite{DBLP:conf/cvpr/WangHcGSWWZ19, DBLP:conf/nips/HeHWYASW21, DBLP:conf/cvpr/0004ZZLJCL20}.
However, most of the available datasets could only provide the coarse-grained text-image alignment signals \cite{DBLP:conf/nips/HeHWYASW21}, i.e., instruction-level correspondences to the complete trajectories, where finer ones are required to learn the cross-modality alignment for well-performed navigation \cite{DBLP:conf/emnlp/HongOWG20, DBLP:conf/acl/ZhuHCDJIS20, DBLP:conf/iccv/QiPH0H021, DBLP:conf/nips/MoudgilMALB21,  DBLP:conf/cvpr/WangLSGW22}.

A large part of previous works \cite{DBLP:conf/cvpr/Zhu0CL20, DBLP:conf/nips/FriedHCRAMBSKD18, DBLP:conf/eccv/MajumdarSLAPB20, DBLP:journals/corr/abs-2203-11591} has concentrated on the global alignment of instructions and trajectories, which matches instructions to the overall temporal visual trajectory via reinforcement learning \cite{DBLP:conf/cvpr/WangHcGSWWZ19}, auxiliary reasoning \cite{DBLP:conf/cvpr/Zhu0CL20}, or transformer-based matching pre-training \cite{DBLP:conf/nips/ChenGSL21, DBLP:journals/corr/abs-2203-11591,DBLP:conf/eccv/MajumdarSLAPB20}.  
Others \cite{DBLP:conf/emnlp/HongOWG20, DBLP:conf/acl/ZhuHCDJIS20, DBLP:conf/nips/HeHWYASW21} have attempted to align sub-instructions and sub-trajectories locally.
At this granularity, the agents are designed to segment long instruction \cite{DBLP:conf/acl/ZhuHCDJIS20,DBLP:conf/emnlp/HongOWG20} and determine which sub-instruction to focus on. 
To guide local cross-modal alignment, He \emph{et al.} \cite{DBLP:conf/nips/HeHWYASW21} introduced the Landmark-RxR dataset by human-annotating sub-instructions and sub-trajectories alignment.
They advanced the granularity of visual-textual alignment, but the agent learning is still supervised at the sentence level.
The alignment within (sub-)instructions should be considered.

Identifying environment landmarks that correspond to entities in the instruction, as shown in Figure \ref{fig:example}, is the next step to refine alignment and improve navigation. 
Recent studies \cite{DBLP:conf/iccv/QiPH0H021, DBLP:conf/nips/MoudgilMALB21, DBLP:conf/cvpr/WangLSGW22} investigated the same problem, suggesting the alignment of entities and object regions \cite{DBLP:conf/iccv/QiPH0H021} or the scene- and object-aware transformer model \cite{DBLP:conf/nips/MoudgilMALB21}.
But the object regions were not under the direct supervision of the corresponding entity phrases in the instruction. 
To generate navigation instructions enriched with landmark phrases, Wang \emph{et al.} \cite{DBLP:conf/cvpr/WangLSGW22} introduced grounded landmark annotations using a dependency parser and weak supervision from the pose traces, leading to the entity-landmark alignments of insufficient precision. 
Therefore, a dataset with high-quality entity-landmark level grounding  annotations and powerful supervision for entity-landmark level cross-modal alignment is highly desired for VLN. 

To address the above limitations, we first enhance the Room-to-Room (R2R) dataset \cite{DBLP:conf/cvpr/AndersonWTB0S0G18} by introducing additional high-quality grounded entity-landmark human annotations, known as the Grounded Entity-Landmark R2R dataset (GEL-R2R). 
The GEL-R2R dataset is annotated with abundant precisely matched entity-landmark pairs (as magenta text and bounding box illustrated in Figure \ref{fig:example}), which could provide the VLN models with direct supervision of fine-grained cross-modal alignment.

To verify the value of our GEL-R2R dataset, we propose a novel Grounded Entity-Landmark Adaptive (GELA) pre-training paradigm to improve the learning of entity-landmark level alignment for the VLN pre-trained model.
Specifically, we suggest three grounded entity-landmark adaptive pre-training objectives:
1) Entity Phrase Prediction (EPP), locating entity phrases that refer to environment landmarks from the instruction;
2) Landmark Bounding box Prediction (LBP), predicting the bounding box of environment landmarks that match with entity phrases;
and 3) Entity-Landmark Semantic Alignment (ELSA), aligning the matched pairs of landmark patches and entity tokens in the feature space by contrastive loss. 
These three tasks explicitly equip the model with the ability to comprehend the entity-level grounding between human instructions and visual environment observations.

Finally, we conduct extensive experiments on two downstream tasks to evaluate our proposed dataset GEL-R2R and adaptive pre-training methods GELA: Room-to-Room (R2R) and Vision-and-Dialog Navigation (CVDN). 
The instructions in R2R are fine-grained descriptions for the navigation trajectory, whereas the instructions in CVDN are multi-turn dialogs between the agent and the oracle during navigation.
We use HAMT \cite{DBLP:conf/nips/ChenGSL21} as the backbone VLN model for the GELA pre-training. 
The results demonstrate our suggested GELA achieves state-of-the-art (SoTA) performance in both seen and unseen environments of the above two benchmarks: 62\% SPL on R2R and 5.87 GP on CVDN.

To summarize, our contributions are three-fold:
\begin{compactitem}
    \item We construct a new dataset GEL-R2R, which is the first dataset with high-quality grounded entity-landmark human annotations in the VLN domain. 
    \item We propose a novel Grounded Entity-Landmark Adaptive (GELA) pre-training paradigm for VLN, explicitly supervising the models to learn fine-grained cross-modal semantic alignment between entity phrases and environment landmarks. 
    \item Our suggested GELA achieves state-of-the-art results on two challenging VLN downstream benchmarks, demonstrating its effectiveness and generalizability.
\end{compactitem}

\section{Related Work}
\label{sec:Related Work}
\textbf{Vision-and-Language Navigation.}
Since the R2R benchmark was proposed by \cite{DBLP:conf/cvpr/AndersonWTB0S0G18},  studies of the VLN have made considerable progress in various aspects. 
VLN datasets have become increasingly diverse, existing in indoor \cite{DBLP:conf/3dim/ChangDFHNSSZZ17,DBLP:conf/emnlp/KuAPIB20, DBLP:conf/cvpr/QiW0WWSH20, DBLP:conf/corl/ThomasonMCZ19, DBLP:conf/cvpr/ZhuL0YCL21} or outdoor \cite{DBLP:conf/cvpr/ChenSMSA19, DBLP:conf/aaai/HermannMMBAH20, DBLP:conf/emnlp/MisraBBNSA18} environments, and in discrete \cite{DBLP:conf/cvpr/AndersonWTB0S0G18, DBLP:conf/cvpr/QiW0WWSH20} or continuous \cite{DBLP:conf/eccv/KrantzWMBL20, DBLP:conf/eccv/KrantzL22, DBLP:journals/ral/LiGLS22, DBLP:conf/cvpr/HongWWG22} environments. 
The architecture of VLN models has become more complex, progressing from LSTM-based \cite{DBLP:conf/nips/FriedHCRAMBSKD18, DBLP:conf/iclr/MaLWAKSX19} to transformer-based \cite{DBLP:conf/cvpr/HaoLLCG20,DBLP:conf/nips/MoudgilMALB21, DBLP:conf/iccv/PashevichS021, DBLP:conf/nips/ChenGSL21}.  
To develop the competence of getting closer to the destination and to the ground truth path \cite{DBLP:conf/acl/JainMKVIB19}, VLN agents have been enhanced mainly through action strategy learning \cite{DBLP:conf/ijcai/ZhangTB20, DBLP:conf/cvpr/ChenGTSL22, DBLP:conf/eccv/QiPZHW20, DBLP:conf/nips/HongOQ0G20, DBLP:conf/iccv/HuangJMKMBI19} and multi-modal representation learning  \cite{DBLP:conf/nips/IlharcoJKIB19, DBLP:conf/eccv/WangWSLS20, DBLP:conf/iccv/KohLYBA21, DBLP:conf/cvpr/KeLBHGLGCS19, DBLP:conf/mm/AnQHWWT21}. 
Many excellent VLN models use imitation and reinforcement learning-based training paradigms for action strategy learning \cite{DBLP:conf/cvpr/WangHcGSWWZ19, DBLP:conf/cvpr/NguyenDBD19, DBLP:conf/naacl/TanYB19}. 
Recently, multi-modal transformer-based models that use effective joint representations for instructions and visual observations have achieved promising performance in VLN \cite{DBLP:conf/iccv/GuhurTCLS21, DBLP:conf/nips/ChenGSL21, DBLP:journals/corr/abs-2203-11591}.
In this passage, we focus on incorporating more accurate and fine-grained cross-modal grounding information to improve the performance of VLN models by supervising the alignment between entity phrases and environment landmarks. 

\begin{table*}[tb]
\begin{center}
  \resizebox{1\textwidth}{!}{
  \begin{tabular}{lcccccc|ccc|ccc}
    \toprule
     & \multicolumn{6}{c}{All}  &\multicolumn{3}{c}{Object} & \multicolumn{3}{c}{Scene} \\
      \cmidrule{2-13}
       & Trajectory & Instruction & Phrase & P/I  & Box    & B/I  & Phrase   & Box     & P/B   & Phrase  & Box & P/B  \\
      \midrule
Train             & 4675       & 14039       & 57788  & 4.12 & 121146 & 8.63 & 30756    & 63248   & 2.06  & 27032   & 57902    & 2.14 \\
Val Seen          & 340        & 1021        & 4196   & 4.11 & 8741   & 8.56 & 1939     & 4065    & 2.10  & 2257    & 4676     & 2.07 \\
Val Unseen        & 775        & 2325        & 9483   & 4.08 & 20296  & 8.73 & 4324     & 8965    & 2.07  & 5159    & 11331    & 2.20 \\
  \midrule
Total             & 5790       & 17385       & 71467  & 4.11 & 150183 & 8.64 & 37019    & 76278   & 2.06  & 34448   & 73909    & 2.15 \\
  \bottomrule
  \end{tabular}
  }
\end{center}
  \caption{
  Statistics on our GEL-R2R dataset. P/I (resp. B/I) denotes the average number of entity phrases contained (resp. landmark bounding boxes matched with entity phrases) in each instruction. And P/B represents the average number of landmark bounding boxes matched with each entity phrase.
  }
  \label{table:dataset}
\end{table*}

\textbf{Phrase-to-Region Grounding.}
Phrase-to-region grounding is an important task in the Vision-Language (VL) domain that involves localizing textual entities in an image, commonly abbreviated as \emph{phrase grounding} \cite{DBLP:conf/iccv/KamathSLSMC21, DBLP:conf/cvpr/LiZZYLZWYZHCG22, lu-etal-2022-extending}.  
This task has seen significant progress since the introduction of the Flickr30k Entities dataset \cite{DBLP:journals/ijcv/PlummerWCCHL17} and has played a crucial role in learning fine-grained semantic visual representation. 
Various models such as VisualBERT \cite{DBLP:journals/corr/abs-1908-03557}, MDETR \cite{DBLP:conf/iccv/KamathSLSMC21}, and GLIP \cite{DBLP:conf/cvpr/LiZZYLZWYZHCG22} have explored this task under different architectures, and grounded pre-training has shown to facilitate fine-grained semantic understanding. 
However, the task of fine-grained cross-modal semantic alignment in embodied VLN tasks is more challenging due to more complex scenes and various objects. 
In this paper, we introduce high-quality entity-landmark grounding human annotations into the R2R dataset and propose a grounded entity-landmark adaptive pre-training scheme for VLN pre-trained models to address this problem. 

\textbf{Adaptive Pre-training.}
In the field of natural language processing, a pre-trained language model in the general domain is continuously pre-trained to learn the knowledge appropriate for a particular task or domain. 
This process is known as adaptive pre-training \cite{DBLP:conf/acl/GururanganMSLBD20}, which is conducted between pre-training and fine-tuning.  
Previous work \cite{DBLP:journals/corr/abs-2009-13570, DBLP:conf/emnlp/ZhaoMZ0W21} gains consistent improvements by continuously pre-training an adaptive language model with the Masked Language Model (MLM) \cite{DBLP:conf/naacl/DevlinCLT19} objective. 
Furthermore, through three pre-training objectives — the MLM, Span Boundary Objective (SBO) \cite{DBLP:journals/tacl/JoshiCLWZL20}, and Perturbation Masking Objective (PMO) — Wu \emph{et al.} \cite{DBLP:conf/acl/WuXSJZS20} improved the overall performance of a dialogue understanding model. 
In this work, we adopt adaptive pre-training into the area of VLN. 
Specifically, based on our human-annotated GEL-R2R dataset, we continuously train the SoTA pre-trained model \cite{DBLP:conf/nips/ChenGSL21} with three different supervised entity-landmark grounding objectives.

\begin{figure*}\centering
\includegraphics[scale=0.75]{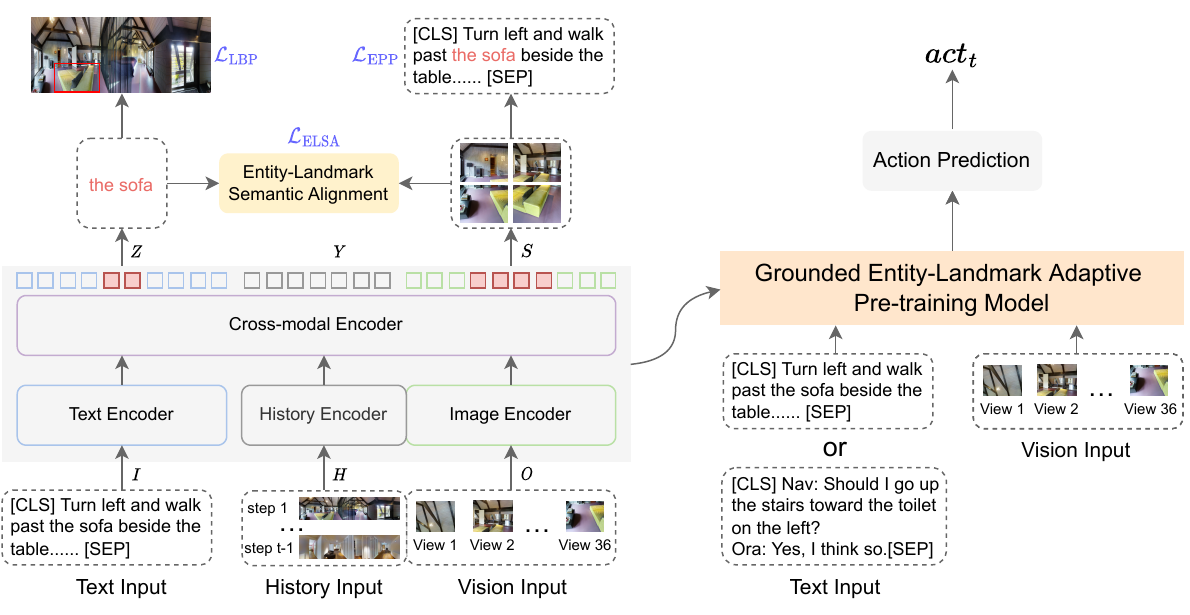}
\caption{Overview of GELA. The three adaptive pre-training objectives served by our scheme encourage: 1) the annotated entity to predict the bounding box of its corresponding landmark in panoramas, 2) the annotated landmark to predict the positions of its corresponding entity phrases, and 3) the cross-modal encoder to establish the semantic alignment between landmarks and entities representations. After grounded entity-landmark adaptive pre-training, the GELA model is fine-tuned for the two VLN downstream tasks, R2R and CVDN.}
\label{fig:model}
\end{figure*}

\section{GEL-R2R Dataset}
\subsection{Dataset Construction}

To establish alignment between entity phrases in instructions and their corresponding landmarks in the surrounding environment, we introduce GEL-R2R, a dataset that includes human annotations of grounded entity-landmark pairs. The construction process consists of five stages: 

\noindent (1) Raw data preparation. 
We collect panoramas of the R2R dataset from the Matterport3D (MP3D) simulator \cite{DBLP:conf/3dim/ChangDFHNSSZZ17} for each viewpoint. 
To improve the accuracy and efficiency of annotating landmarks, we mark the next action direction in panoramas and match each panorama with the corresponding sub-instruction based on \cite{DBLP:conf/emnlp/HongOWG20}. 

\noindent (2) Annotation tool development.
We develop a web-based annotation platform using the label-studio \cite{Label/Studio} to facilitate annotations. 
The annotation interface is shown in the supplementary material. 
Annotators first mark the entity phrases in the instructions and then identify the corresponding landmarks in the images using the same labels.

\noindent (3) Annotation guideline standardization.
Through pre-annotation, we standardized the annotation guideline by establishing four rules to ensure the consistency of the annotations. 
The four rules are as follows:
\begin{compactitem}
     \item \textbf{Alignment Rule}: The entity phrase in instructions should match landmarks in panoramas precisely. 
     \item \textbf{Free Text Rule}: Free text instead of the class should be annotated, for instance, ``the white dining table'' instead of ``table''.
     \item \textbf{Text Coreference Rule}: Entity phrases referring to the same landmark are marked with the same label.
     \item \textbf{Unique Landmark Rule}: For an entity phrase, only one corresponding landmark bounding box should be annotated in a panorama. 
\end{compactitem}

\noindent (4) Data annotation and revision. 
We first select annotators by testing the participants using 50 test instruction-path pairs. 
The qualification process leaves us 43 qualified annotators to complete the annotation task.
Following the annotation, our experts double-check the annotations and correct any errors to ensure compliance with the above four rules. 

\noindent (5) Data processing.
In order to ensure the quality of the annotations, we first eliminate the incorrect annotations violating the above rules and then correct a few incorrect words.
Finally, we established the GEL-R2R dataset by introducing additional grounded entity-landmark annotations to the R2R dataset. 
More details about data collection are provided in the supplementary material.

\subsection{Statistics and Analysis}

Table \ref{table:dataset} presents statistics for the train, validation seen, and validation unseen splits of GEL-R2R. 
The dataset contains a total of 71,467 entity phrases, with 57,788 in the train split, 4,196 in the validation seen split, and 9,483 in the validation unseen split. 
It also includes 150,183 landmark boxes, of which 121,146 are in the train split, 8,741 are in the validation seen split, and 20,296 are in the validation unseen split.
On average, each instruction contains 4.1 entity phrases, which is roughly equal to the number of actions in a path. Furthermore, each instruction uses about 8.5 landmark boxes on average. 
As a result, the agent has access to a diverse set of landmarks that can aid in decision-making during each navigation episode. 

We compute statistics on the object and scene, taking into account their varying identification difficulties. 
In total, we gather 37019 object entity phrases and 34448 scene entity phrases, referring to 76278 object landmark boxes and 73909 scene landmark boxes, respectively.
On average, there are about 2 landmark boxes corresponding to each exact entity phrase, whether for objects or scenes. 
This indicates that the same landmark appears in approximately two adjacent viewpoint panoramas from different sizes and angles, which enables the agent to perceive the landmark from multiple views.
In summary, the GEL-R2R dataset is the first to provide high-quality entity-landmark alignment human annotations, which are essential for the area of VLN, such as cross-modal representation learning, data augmentation, and interpretable navigation.

\section{Method}
\subsection{Problem Setup and Overview}

\textbf{Problem Setup.}
In VLN tasks, an embodied agent should move to a target location from a starting pose inside a 3D photo-realistic environment by following natural language navigable instructions. 
The instruction is a sequence of ${L}$  words, denoted as $I=\left\{w_{0}, \ldots, w_{L-1}\right\}$, and it guides the agent to traverse a connectivity graph ${G}$ in order to reach the intended destination.
At each step ${t}$, the agent acquires a new panoramic visual observation $O_{t}=\left\{o_{t, i}\right\}_{i=1}^{36}$ from neighboring environment, where $o_{t, i}=\left[v_{t, i} ; a_{t, i}\right]$ consists RGB image $v_{t, i}$ and orientation angle  $a_{t, i}$ of ${i}$-th view. 
The relative angles of ${n}$ navigable viewpoints to the current viewpoint and the stop action constitute the action space at step ${t}$, denoted as $A_{t}=\left\{act_{t, 1}, \ldots, act_{t, n}, [STOP]\right\}$. 
The agent selects an action to move to a navigable viewpoint in ${G}$ or stop at the current location. 
Once the agent stops, the navigation episode is completed. 

\textbf{Overview.}
In this study, we propose a novel grounded entity-landmark adaptive pre-training scheme for a VLN pre-trained model, whose architecture is shown in Figure \ref{fig:model}. 
The training scheme introduces explicit supervision on entity-landmark grounding to enhance fine-grained cross-modal representation learning.
The scheme comprises two distinct stages: GELA pre-training on several proxy objectives and fine-tuning on the VLN downstream tasks. 

\subsection{Pre-trained Model}
As illustrated in the bottom-left of Figure \ref{fig:model}, our pre-trained model is a fully transformer-based architecture for multi-modal decision-making, which is modified from the classical cross-modal model LXMERT \cite{DBLP:conf/emnlp/TanB19}. 
The pre-trained model takes three inputs: a global instruction ${I}$, history information $H_{t}$, and current panoramic visual observation $O_{t}$, which are fed in a language encoder, a history encoder, and a vision encoder, respectively. 
And then the textual and visual modalities exchange the signals through cross-attention layers in the cross-modal encoder. 
Specifically, the visual modality is the concatenation of history and visual observation.
Finally, the representation of tokens in instruction, history, and visual state is 
$Z=\left\{z_{\text{cls}}, z_1, \cdots, z_{L}\right\}$, $Y_t=\left\{y_{\text{cls}},y_1, \cdots, y_{t-1}\right\}$, and $S_t=\left\{s_{1}, \cdots, s_{36}, s_{\text {stop }}\right\}$, respectively.

To learn effective uni-modal and multi-modal representation, transformer-based models for VLN commonly undergo pre-training on in-domain datasets using several proxy tasks  \cite{DBLP:conf/cvpr/HaoLLCG20, DBLP:journals/corr/abs-2203-11591, DBLP:conf/nips/ChenGSL21}.  These tasks include common vision-language pre-training tasks as well as VLN-specific auxiliary tasks. 
In this study, we utilize five proxy tasks: 
Masked Language Modeling (MLM) \cite{DBLP:conf/naacl/DevlinCLT19}, 
Masked Region Classification (MRC) \cite{DBLP:conf/nips/LuBPL19}, 
Instruction Trajectory Matching (ITM) \cite{DBLP:conf/eccv/MajumdarSLAPB20}, 
Single-step Action Prediction (SAP) \cite{DBLP:conf/nips/ChenGSL21},  
and Spatial Relationship Prediction (SPREL) \cite{DBLP:conf/nips/ChenGSL21}. 
Details are demonstrated in the supplementary material. 

\subsection{GELA Pre-training}
As shown in the top-left of Figure \ref{fig:model}, we adopt three adaptive pre-training objectives for learning entity-landmark level alignment, which are Entity Phrase Prediction (EPP), Landmark Bounding box Prediction (LBP), and Entity-Landmark Semantic Alignment (ELSA). 
The three proxy objectives are designed by mimicking the objectives used in pre-training vision-language Transformers, particularly with the goal of learning phase-level grounding information. 
LBP is adapted from the training objective of TransVG \cite{DBLP:conf/iccv/DengYCZL21}.
EPP and ELSA are extended from the soft token prediction and contrastive alignment of MDETR \cite{DBLP:conf/iccv/KamathSLSMC21}, respectively.

\textbf{Entity Phrase Prediction (EPP).}
In this objective, we predict the positions of entity phrases that correspond to annotated environment landmarks. 
We first transfer the human-annotated entity location to a mask vector $M_z$ of ${L}$+1 dimensions, which is the same as $Z$. 
Similarly, we transfer the human-annotated landmark bounding box to a 37-dim (the same as $S_t$) mask vector $M_s$. 
Then we train the model to infer a uniform distribution over all token positions that refer to the corresponding landmark patches and supervise this process by the mask vector $M_z$. 
Specifically, we average the patches representation of the landmark: $S_t {\times} M_{s}^{\top}$, where ${\times}$ denotes the matrix product, obtaining a 768-dim vector.  
Then we put it into a two-layer feedforward network (FFN) to predict a distribution over the token positions in the instruction sequence: 
\begin{equation}
\label{eq:logits}
    logits = \operatorname{Softmax} (\operatorname{FFN} (S_t {\times} M_{s}^{\top}) ),
\end{equation}
where the FFN maps a vector of 768-dim to a vector of $L$+1 dimensions. 
Finally, we minimize the cross-entropy loss between $logits$ and $M_{z}$: 
\begin{equation}
\label{eq:lce}
    \mathcal{L}_\text{EPP} = \operatorname{CrossEntropy} (logits, M_{z}).
\end{equation}

\textbf{Landmark Bounding Box Prediction (LBP).}
In this objective, we predict the bounding box of the landmark that matches with annotated entity phrases.  
We train the model to directly predict a 4-dim vector  $box^{\prime} = (x^{\prime}, y^{\prime}, w^{\prime}, h^{\prime})$ as the coordinates of the bounding box for each entity phrases and supervise this process by the human-annotated bounding box  $box = (x, y, w, h)$. 
Specifically, we first average the token embeddings of the entity phrase: $Z {\times} M_{z}^{\top}$ and then we predict the coordinates of the box through a two-layer FFN and sigmoid function: 
\begin{equation}
\label{eq:box}
    box^{\prime} = \operatorname{Sigmoid}( \operatorname{FFN} (Z {\times} M_{z}^{\top})),
\end{equation}
where the FFN maps a vector of 768-dim to a vector of 4-dim. 
Finally, we apply the smooth L1 loss and generalized IoU loss (GIoU loss) to optimize the box coordinates $M_{z}$:
\begin{equation}
\label{eq:lbox}
    \mathcal{L}_\text{LBP} = \mathcal{L}_{\text{smooth-}l_{1}}(box,box^{\prime})+{\lambda}\mathcal{L}_\text{GIoU}(box,box^{\prime}),
\end{equation}
where $\lambda$ is the weight coefficient of GIoU loss to balance these two losses.

\textbf{Entity-Landmark Semantic Alignment (ELSA).}
While the above two unidirectional prediction tasks use positional information to match the entity and landmark, the entity-landmark semantic alignment loss enforces alignment between the hidden embeddings of the landmark and entity at the output of the cross-modal encoder. 
This additional contrastive alignment loss ensures that the representations of the landmark patches and the corresponding entity tokens are closer in the feature space compared to representations of unrelated tokens. 
This constraint is stronger than the above two unidirectional prediction losses as it directly operates on the representations and is not single-handedly based on the positional information. 
Specifically, inspired by InfoNCE loss \cite{DBLP:journals/ijon/LiuMXYTPG22}, the objective is the mean of two contrastive losses as follows:

\begin{equation}
\label{eq:ls}
\fontsize{9.5pt}{8pt}\selectfont
    \mathcal{L}_s=\sum_{i=0}^{36} \frac{1}{\left|Z_i^{+}\right|} \sum_{j \in Z_i^{+}}-\log \left(\frac{\exp \left(s_i^{\top} z_j / \tau\right)}{\sum_{k=0}^{L} \exp \left(s_i^{\top} z_k / \tau\right)}\right),
\end{equation}

\begin{equation}
\label{eq:lz}
\fontsize{9.5pt}{8pt}\selectfont
    \mathcal{L}_z=\sum_{i=0}^{L} \frac{1}{\left|S_i^{+}\right|} \sum_{j \in s_i^{+}}-\log \left(\frac{\exp \left(z_i^{\top} s_j / \tau\right)}{\sum_{k=0}^{36} \exp \left(z_i^{\top} s_k / \tau\right)}\right),
\end{equation}

\begin{equation}
\label{eq:lcon}
    \mathcal{L}_\text{ELSA} = (\mathcal{L}_s+\mathcal{L}_z)/2,
\end{equation}
where $Z_i^{+}$ is the token set to be matched with an annotated patch $s_i$, and $S_i^{+}$ is the landmark patch set that should be matched with an annotated token $z_i$, and $\tau$ is a temperature parameter regulating attention to negative samples.

Therefore, the full adaptive pre-training objective is:
\begin{equation}
\label{eq:lgela}
    \mathcal{L}_\text{GELA} = \alpha\mathcal{L}_\text{EPP}+\beta\mathcal{L}_\text{LBP}+\gamma\mathcal{L}_\text{ELSA}.
\end{equation}

\subsection{Fine-tuning for VLN Tasks}
\label{sub:ft}
The fine-tuning stage is illustrated in the right of Figure \ref{fig:model}, where we generalize the GELA model for two VLN downstream tasks under the scheme of imitation learning (IL) and reinforcement learning (RL) by following previous work \cite{DBLP:conf/cvpr/Hong0QOG21,DBLP:conf/nips/ChenGSL21}.
IL supervises the agent to clone the behavior of the expert while RL encourages the agent to explore the trajectory according to the learning policy. 
Firstly, the GELA model navigates in the environment following the ground-truth action and generates gradients by IL. 
Secondly, using the same instruction, the model samples the action space to make decisions and uses RL to generate gradients. 
Finally, we combine the gradients and optimize the pre-trained model. 
For the CVDN task, our setting is the same to \cite{DBLP:conf/nips/ChenGSL21}. 
For the R2R task, we apply three settings with different data augmentation following \cite{DBLP:conf/cvpr/LiTB22}:
\begin{compactitem}
     \item \texttt{none}: without data augmentation on environments.
     \item \texttt{st}: with data augmentation by style transformation on original environments. Specifically, each discrete view of panoramas is transferred with a random style.
     \item \texttt{smo}: with data augmentation by semantic class masking on synthesis environments, which are generated by image synthesis with the same style as the original environments. One semantic class is randomly masked out during image generation. 
\end{compactitem}

\section{Experiment and Results}
\subsection{Experimental Setup}
\noindent
\textbf{Datasets.}
We evaluate our proposed method on the two VLN datasets: Room-to-Room (R2R) \cite{DBLP:conf/cvpr/AndersonWTB0S0G18} and Vision-and-Dialog Navigation (CVDN) \cite{DBLP:conf/corl/ThomasonMCZ19}, which are all based upon the MP3D indoor environments. 
The instructions in R2R are detailed descriptions of the navigation trajectory.
There are four splits in 21,558 trajectory-instruction pairs of the R2R dataset: training (14,039), validation seen (1,021), validation unseen (2,325), and test unseen (4,173).
Our grounded entity-landmark human annotations are based on the training, validation seen, and validation unseen splits. 
The instructions in CVDN are multi-turn dialogs between the agent and the oracle during navigation.
These kinds of instructions are frequently vague and unspecific.
Therefore, the CVDN task is more challenging, serving to evaluate the generalization ability for new downstream tasks.

\noindent
\textbf{Evaluation Metrics.}
For R2R, we report four evaluation metrics: 
Trajectory Length (TL); 
Navigation Error (NE $\downarrow$) - the average distance in meters between the agent’s final position and the goal viewpoint; 
Success Rate (SR $\uparrow$) - the proportion of paths where the agent stopped within 3 meters of the goal viewpoint; 
Success rate weighted by Path Length (SPL $\uparrow$) \cite{DBLP:journals/corr/abs-1807-06757}
\footnote{
We denote a $\downarrow$ to demonstrate lower is better and an $\uparrow$ to demonstrate higher is better.
}. 
SR and SPL are the recommended key metrics. 
For CVDN, we use the average progress in meters of the agent towards the goal viewpoint as the primary evaluation metric, denoted as Goal Progress (GP $\uparrow$).

\begin{table}[tb]
\begin{center}
\resizebox{0.43\textwidth}{!}{
\begin{tabular}{llcc>{\columncolor[gray]{0.9}}c>{\columncolor[gray]{0.9}}c}
\toprule[1.25pt]
\multirow{2}{*}{Model} & \multirow{2}{*}{Feature} & \multicolumn{4}{c}{Validation Unseen} \\
\cmidrule{3-6}
                       &                          & TL       & NE$\downarrow$      & SR$\uparrow$       & SPL$\uparrow$     \\
                       \midrule
\multirow{3}{*}{HAMT\cite{DBLP:conf/nips/ChenGSL21}}  & \texttt{none}                     & 11.46    & \textbf{2.29}    & 65.7    & 60.9   \\
                       & \texttt{st}                      & 11.78    & 3.42    & 67.3    & 62.6   \\
                       & \texttt{smo}                     & 12.13    & 3.22    & 67.9    & 62.9   \\
                       \midrule
\multirow{3}{*}{GELA (ours)}  & \texttt{none}                     & 11.75    & 3.33    & 69.2    & 63.4   \\
                       & \texttt{st}                      & 11.56    & 3.26    & 69.3    & 64.2   \\
                       & \texttt{smo}                     & 11.73    & 3.11    & \textbf{71.1}    & \textbf{65.0} \\
                       \bottomrule[1.25pt]
\end{tabular}
}
\end{center}
  \caption{
  Comparison with the HAMT baseline on the R2R validation unseen split. \textbf{Black} denotes the best results.
  }
  \label{table:gela}
\end{table}

\begin{table*}[tb]
\begin{center}
  \resizebox{0.9\textwidth}{!}{
\begin{tabular}{lcc>{\columncolor[gray]{0.9}}c>{\columncolor[gray]{0.9}}ccc>{\columncolor[gray]{0.9}}c>{\columncolor[gray]{0.9}}ccc>{\columncolor[gray]{0.9}}c>{\columncolor[gray]{0.9}}c}
\toprule[1.25pt]
\multicolumn{1}{l}{\multirow{2}{*}{Methods}} & \multicolumn{4}{c}{Validation Seen} & \multicolumn{4}{c}{Validation Unseen} & \multicolumn{4}{c}{Test Unseen} \\
\cmidrule{2-13} 
   & TL       & NE$\downarrow$     & SR$\uparrow$     & SPL$\uparrow$    & TL       & NE$\downarrow$      & SR$\uparrow$      & SPL$\uparrow$    & TL      & NE$\downarrow$    & SR$\uparrow$   & SPL$\uparrow$   \\
\midrule
Seq2Seq\cite{DBLP:conf/cvpr/AndersonWTB0S0G18}                  & 11.33    & 6.01   & 39     & -      & 8.39     & 7.81    & 22      & -      & 8.13    & 7.85  & 20    & 18    \\
$\text{SSM}$ \cite{DBLP:conf/cvpr/WangWLXS21}                      & 14.70    & 3.10   & 71   & 62   & 20.70    & 4.32    & 62    & 45   & 20.40   & 4.57  & 61  & 46  \\
EnvDrop\cite{DBLP:conf/naacl/TanYB19}                  & 11.00    & 3.99   & 62     & 59     & 10.70    & 5.22    & 52      & 48     & 11.66   & 5.23  & 51    & 47    \\
AuxRN\cite{DBLP:conf/cvpr/Zhu0CL20}                    & -        & 3.33   & 70     & 67     & -        & 5.28    & 55      & 50     & -       & 5.15  & 55    & 51    \\
SEvol\cite{DBLP:conf/cvpr/ChenGMZ022}                    & 11.97    & 3.56   & 67     & 63     & 12.26    & 3.99    & 62      & 57     & 13.40   & 4.13  & 62    & 57    \\
\midrule
PREVALENT\cite{DBLP:conf/cvpr/HaoLLCG20}                & 10.32    & 3.67   & 69     & 65     & 10.19    & 4.71    & 58      & 53     & 10.51   & 5.30  & 54    & 51    \\
AirBERT\cite{DBLP:conf/iccv/GuhurTCLS21}                  & 11.09    & 2.68   & 75     & 70     & 11.78    & 4.01    & 62      & 56     & 12.41   & 4.13  & 62    & 57    \\
RecBERT\cite{DBLP:conf/cvpr/Hong0QOG21}                  & 11.13    & 2.90   & 72     & 68     & 12.01    & 3.93    & 63      & 57     & 12.35   & 4.09  & 63    & 57    \\
HOP\cite{DBLP:journals/corr/abs-2203-11591}                      & 11.26    & 2.72   & 75     & 70     & 12.27    & 3.80    & 64      & 57     & 12.68   & 3.83  & 64    & 59    \\
$\text{REM}$\cite{DBLP:conf/iccv/0002ZCLGS21}                     & 10.88        & 2.48      & 75      &  72      & 12.44    & 3.89    &64      & 58     &  13.11   &  3.87  & 65    & 59  \\
$\text{HAMT}$\cite{DBLP:conf/nips/ChenGSL21}                     & 11.15    & 2.51   & \textbf{76}     & 72     & 11.46    & \textbf{2.29}    & 66      & 61     & 12.27   & 3.93  & 65    & 60    \\
\midrule
$\text{GELA (ours)}$               & 11.19    & \textbf{2.39}   & \textbf{76}     & \textbf{73}     & 11.73    & 3.11    & \textbf{71}      & \textbf{65}     & 12.99   & \textbf{3.59}  & \textbf{67}    & \textbf{62} \\
\bottomrule[1.25pt]

\end{tabular}
}
\end{center}
  \caption{
  Comparison with SoTA methods on the R2R dataset. The methods in the top group are trained from scratch. The methods in the second group are based on pre-training. 
  }
  \label{table:r2r}
\end{table*}

\noindent
\textbf{Implementation Details.} 
We typically conform to the architecture of \cite{DBLP:conf/nips/ChenGSL21} and its associated hyper-parameters. 
We conduct adaptive pre-training on our GEL-R2R training dataset and the augmented dataset from \cite{DBLP:conf/cvpr/HaoLLCG20}. 
In the adaptive pre-training stage, we inherited the pre-training tasks to prevent knowledge forgetting.
The GEL-R2R training set serves for both pre-training tasks and adaptive pre-training tasks.
Due to the lack of entity-landmark grounding annotations, the augmented dataset serves only for pre-training tasks.
We set $\lambda$=1.0 in Eq. (\ref{eq:lbox}), and $\alpha$=1.0, $\beta$=1.0, $\gamma$=1.0 in Eq. (\ref{eq:lgela}). 
And we set $\tau$ to 0.07 in Eq. (\ref{eq:ls}) and Eq. (\ref{eq:lz}) following \cite{DBLP:conf/iccv/KamathSLSMC21}.  
And We adopt adaptive pre-training for 200k iterations with a learning rate of 5e-5 by using 2 NVIDIA RTX 3090 GPUs, and the batch size for each GPU is set to 64.
We adopt fine-tuning on the R2R and CVDN tasks respectively. 
For both R2R and CVDN tasks, we train the GELA model for 100k iterations with a learning rate of 1e-5 on a single NVIDIA RTX 3090 GPU. 
We set the batch size to 16 for the R2R task and 8 for the CVDN task.

\subsection{The Effect of GELA Pre-training}
\label{pa:gela}
In this section, we demonstrate the effects of our GELA pre-training paradigm for VLN models in terms of navigation performance and cross-modal representation. 

\begin{figure}[t]
\begin{center}
  \includegraphics[width=0.48\textwidth]{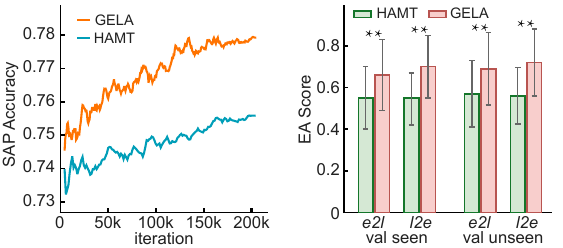}
\end{center}
   \caption{The effect of GELA pre-training. Single-step Action Prediction (SAP) accuracy (left) and Effective Attention (EA) score (right) of GELA and HAMT on the R2R validation splits.}
   \label{fig:eff}
\end{figure}

\noindent
\textbf{The effect of GELA pre-training on navigation performance.}
Table \ref{table:gela} presents a comparison with the HAMT baseline model on the R2R validation unseen split under three fine-tuning settings. 
The results of the first three lines are obtained from \cite{DBLP:conf/cvpr/LiTB22}.
The results show that our GELA model outperforms HAMT by a large margin under all three settings. 
Specifically, GELA achieves absolute improvements of 3.5\%, 2\%, and 3.2\% in SR under \texttt{none}, \texttt{st}, and \texttt{smo} respectively. 
Furthermore, we show the single-step action prediction (SAP) accuracy of GELA and HAMT on the R2R validation unseen split during adaptive pre-training in the left panel of Figure~\ref{fig:eff}. 
As the training progresses, we observe a gradual increase in the performance gap between GELA and HAMT, with GELA and HAMT achieving prediction accuracies of approximately 78\% and 75\%, respectively. 
These significant improvements demonstrate that fine-grained cross-modal alignment between entity phrases and environmental landmarks is highly effective in enhancing agents' navigation performance.

\noindent
\textbf{The effect of GELA pre-training on cross-modal representation.}
To directly demonstrate the effect of GELA pre-training on cross-modal semantic alignment, we analyze cross-modal attention weights at the last Transformer layer during inference.
First, we visualize the attention weights normalized to a scale of 0 to 1 in a heatmap.
We then utilize the annotations in GEL-R2R to mark tokens, which are involved in the grounded entity-landmark pairs, and calculate the attention weights received by each token from the corresponding tokens in the other modality, defined as the Effective Attention (EA) score. 
In the right panel of Figure~\ref{fig:eff}, we compare the average EA score of each marked token in GELA and HAMT on the R2R validation seen and unseen splits, where \textit{e2l} (resp. \textit{l2e}) represents the attention of the entity token to the landmark token (resp. the attention of the landmark token to the entity token). 
Among all comparisons, GELA achieves significantly higher EA scores than HAMT (with a p-value in the two-sample t-test approaching 0), indicating that the grounded entity-landmark pairs have a stronger correspondence in GELA. 
While HAMT has a strong ability for uni-modal representation based on the large-scale transformer and end-to-end pre-training, it can be confused about matching the mentioned landmarks with visual observations. 
The explicit supervision for entity-landmark level alignment in GELA can improve the cross-modal representation and substantially enhance the agent's decision-making ability during navigation. 
Qualitative examples are provided in the supplementary material. 

\subsection{Comparison to SoTA}

\noindent
\textbf{Room-to-Room: R2R.}
Table \ref{table:r2r} reports the performance comparison of various models on the R2R benchmark. 
The ensemble models, SE-Mixed \cite{DBLP:journals/corr/abs-2111-14267} and EnvEdit \cite{DBLP:conf/cvpr/LiTB22}, are excluded for a fair comparison. 
Our proposed GELA model outperforms all the other models in primary metrics (SR and SPL) across all the dataset splits.
This indicates that GELA agents can navigate more accurately and efficiently than the other models. 
Specifically, on the validation seen split, GELA performs comparably to the SoTA HAMT. However, on the validation unseen and test unseen splits, GELA outperforms HAMT by absolute 5\% and 2\% (SR) and 4\% and 2\% (SPL) improvements, respectively. 
This demonstrates that GELA exhibits better generalization ability to unseen environments, which can be attributed to its stronger capability in capturing the semantic features of landmarks after entity-landmark grounding learning.

\begin{table}[tb]
\begin{center}
  \resizebox{0.48\textwidth}{!}{

    \begin{tabular}{lccc}
    \toprule[1.25pt]
    Methods  & Val Seen & Val Unseen & Test Unseen \\
    \midrule
    Seq2Seq\cite{DBLP:conf/corl/ThomasonMCZ19}           & 5.92     & 2.10       & 2.35        \\
    CMN\cite{DBLP:conf/cvpr/0004ZZLJCL20}               & 7.05     & 2.97       & 2.95        \\
    PREVALENT\cite{DBLP:conf/cvpr/HaoLLCG20}         & -        & 3.15       & 2.44        \\
    VISITRON\cite{DBLP:conf/acl/ShrivastavaGLPT22}          & 5.11     & 3.25       & 3.11        \\
    ORIST\cite{DBLP:conf/eccv/WangJIWKR20}             & -        & 3.55       & 3.15        \\
    HOP\cite{DBLP:journals/corr/abs-2203-11591}               & -        & 4.37       & 3.31        \\
    MT-RCM+EnvAg\cite{DBLP:conf/eccv/WangJIWKR20}      & 5.07     & 4.65       & 3.91        \\
    HAMT\cite{DBLP:conf/nips/ChenGSL21}              & 6.91     & 5.13       & 5.58        \\
    \midrule
    GELA (ours)              & \textbf{8.57}   & \textbf{5.86}       & \textbf{5.87}   \\
    \bottomrule[1.25pt]
    \end{tabular}
  }
\end{center}
  \caption{
  Comparison with SoTA methods on the CVDN dataset.
  }
  \label{table:cvdn}
\end{table}

\noindent
\textbf{Vision-and-Dialog Navigation: CVDN.}
Table~\ref{table:cvdn} presents the results on the CVDN benchmark, which utilizes Goal Progress (GP) in meters as the key performance metric. 
During training, we use a mixture of two types of demonstrations (navigator and oracle) as supervision for the trajectory.  
The results indicate that GELA consistently outperforms the other models on both the validation and test unseen environments.  
Particularly, our model achieves up to 1.66-meter improvement over the SoTA model HAMT on the validation seen split.
Additionally, GELA outperforms the SoTA results by 0.73 meters and 0.29 meters on validation unseen and test splits, respectively. 
These results clearly demonstrate the GELA model is effective and generalizable to different types of instructions in more challenging VLN downstream tasks. \\

\begin{table}[tb]
\begin{center}
\resizebox{0.48\textwidth}{!}{
\begin{tabular}{clcc>{\columncolor[gray]{0.9}}c>{\columncolor[gray]{0.9}}c}
\toprule[1.25pt]
Methods & Task         & TL    & NE$\downarrow$   & SR$\uparrow$   & SPL$\uparrow$  \\
\midrule
0     & None         & 11.46 & \textbf{2.29} & 65.7 & 60.9 \\
\midrule
1     & EPP          & 11.83 & 3.63 & 65.8 & 60.7 \\
2     & LBP          & 12.71 & 3.71 & 65.6 & 60.1 \\
3     & ELSA         & 11.73 & 3.58 & 66.3 & 61.7 \\
4     & EPP+LBP      & 12.10 & 3.59 & 67.0 & 61.4 \\
5     & EPP+LBP+ELSA & 11.75 & 3.33 & \textbf{69.2} & \textbf{63.4}\\
\bottomrule[1.25pt]
\end{tabular}
}
\end{center}
  \caption{
  Ablation study of adaptive pre-training objectives on the R2R validation unseen split.
  }
  \label{table:task}
\end{table}

\subsection{Ablation studies}
\noindent
\textbf{Adaptive Pre-training Objective.}
To evaluate the effectiveness of different adaptive pre-training objectives, we conduct an ablation study on R2R validation unseen split under \texttt{none} setting. 
The results are presented in Table \ref{table:task}. 
Model 0 represents the HAMT baseline, while Models 1-3 show the results of combining the original proxy tasks in \cite{DBLP:conf/nips/ChenGSL21} with EPP, LBP, and ELSA, respectively. 
The results demonstrate that ELSA can further improve navigation performance, while the other two unidirectional prediction objectives solely based on positional information do not enhance the pre-trained model. 
This indicates that ELSA provides more effective supervision of cross-modal alignment by directly operating on the feature representations. 
Subsequently, we evaluate the effect of combining EPP and LBP. 
The results show that the combined performance (Model 4) is much better than the separate performance of Model 1 and Model 2, demonstrating that these two objectives are complementary.
Finally, we combined all three objectives, and the results show that this performance (Model 5) improves even further. 
These findings highlight the importance of effective cross-modal alignment and the complementary nature of different adaptive pre-training objectives in improving navigation performance.

\noindent
\textbf{Landmark Category.}
Table \ref{table:class} presents our investigation into the impact of two category entities (object and scene) from GEL-R2R under \texttt{none} setting. 
Our results demonstrate that integrating both object and scene entities can enhance navigation performance. 
Of the two entity types, object entities have a greater influence on performance, as agents find it easier to comprehend individual objects compared to complex scenes. 
Furthermore, the best performance is achieved by combining both entity types, enabling agents to navigate more precisely by utilizing both object and scene landmarks. 
This improvement can be attributed to the fact that the amount of available data for each entity type is only half that of the complete dataset. 
Therefore, it is plausible that the performance of GELA has not reached saturation due to the limited volume of available data. 

\begin{table}[tb]
\begin{center}
\resizebox{0.33\textwidth}{!}{
\begin{tabular}{lcc>{\columncolor[gray]{0.9}}c>{\columncolor[gray]{0.9}}c}
\toprule[1.25pt]
Category  & TL    & NE$\downarrow$   & SR$\uparrow$   & SPL$\uparrow$  \\
\midrule
None   & 11.46 & \textbf{2.29} & 65.7 & 60.9 \\
\midrule
Scene  & 11.84 & 3.44 & 66.7 & 61.2 \\
Object & 11.23 & 3.54 & 66.8 & 61.9 \\
All    & 11.75 & 3.33 & \textbf{69.2} & \textbf{63.4} \\
\bottomrule[1.25pt]
\end{tabular}
}
\end{center}
  \caption{
  Ablation study of landmark categories on the R2R validation unseen split.
  }
  \label{table:class}
\end{table}

\subsection{Limitations and Future Work}
GELA is a preliminary investigation into cross-modal alignment at the entity-landmark level in VLN based on our human-annotated GEL-R2R dataset. 
Due to the high cost of human annotation, the volume of GEL-R2R is relatively limited.
In the future, we plan to expand the dataset using two cost-effective approaches: 1) developing a data augmentation model by using GEL-R2R to produce grounded entity-landmark annotations, and 2) incorporating large-scale \textit{phrase grounding} datasets in the VL domain. 
Additionally, we intend to explore interpretable navigation based on our introduced dataset and method.

\section{Conclusion}

A well-performed VLN agent should have powerful competency in fine-grained cross-modal semantic alignment between entities and landmarks. 
In this paper, we introduce the manually annotated grounded entity-landmark dataset GEL-R2R, which provides powerful cross-modal alignment for VLN at the entity-landmark level. 
We then adopt three grounded entity-landmark adaptive pre-training objectives based on GEL-R2R to facilitate cross-modal semantic alignment learning under explicit supervision. Comprehensive experimental results on two downstream VLN tasks, R2R and CVDN, demonstrate the effectiveness and generalizability of our proposed model, GELA.

\section{Acknowlegement}
We thank ICCV reviewers for their constructive suggestions. 
This work is supported in part by the National Natural Science Foundation of China under Grant 62076250, Grant 61703407, and Grant 61901505.

{\small
\bibliographystyle{ieee_fullname}
\bibliography{egbib}
}

\appendix
\section*{Appendix}

In this appendix, we first present additional details for the collection of GEL-R2R in Sec. \ref{A}. 
And then we provide the implementation details of the pre-trained model in Sec. \ref{B}. 
Finally, we compare several qualitative examples of our GELA model and the HAMT \cite{DBLP:conf/nips/ChenGSL21} baseline in Sec. \ref{C}. 

\section{Data Collection Pipeline}
\label{A}

\subsection{Raw Data Preparation.}
Our grounded entity-landmark annotations are based on the Room-to-Room (R2R) \cite{DBLP:conf/cvpr/AndersonWTB0S0G18} dataset. 
For each navigation trajectory from R2R, we collect a sequence of 360-degree panoramas with an image size of 2048×1024 from Matterport3D simulator \cite{DBLP:conf/3dim/ChangDFHNSSZZ17}. 
Additionally, we employ two skills in the preparation of raw data to increase the effectiveness and efficiency of entity-landmark grounding annotation. 
The first skill is turning each panorama so that the direction of the subsequent action heading is in the center of the image and marking the direction using the red arrow, as shown in Figure \ref{fig:anno}. 
Due to the fact that most landmarks are located near the direction of the next action, this skill might improve the speed with which the annotators identify the landmarks and reduce the phenomenon whereby landmarks are divided by the edge. 
The second skill is utilizing alignment information of sub-instructions and sub-trajectories modified from FG-R2R \cite{DBLP:conf/emnlp/HongOWG20}, where a sub-instruction may sometimes incorrectly match a viewpoint rather than a sub-trajectory. 
As presented in Figure \ref{fig:anno}, ``Turn left and exit out the door beside the TV to the left'' is the first sub-instruction of the overall instruction, and the following two panoramas are the visual observations of the agent at two viewpoints of the corresponding sub-trajectory. 
As a result, rather than having to search through every panorama along the path, annotators only need to find effective landmarks in the several corresponding panoramas.

\subsection{Annotation Tool Development.}
\begin{figure}[tb]
  \centering
  \includegraphics[width=0.48\textwidth]{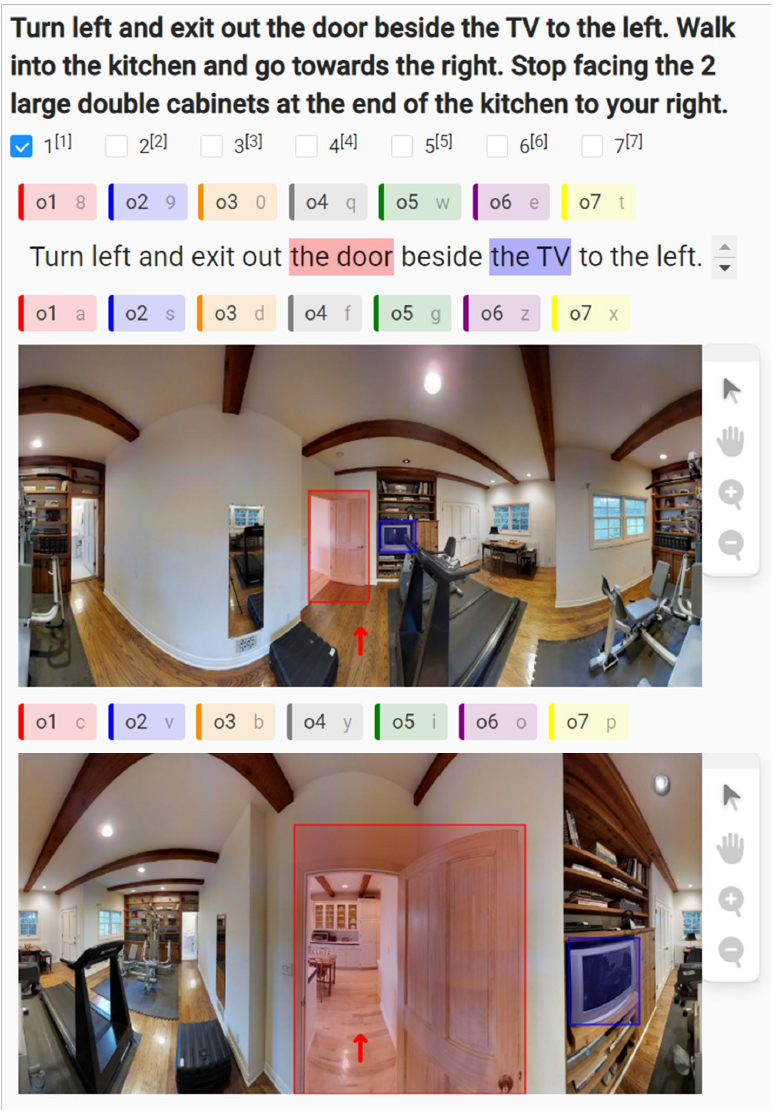}
  \caption{The designed interface for entity-landmark grounding annotation. \textbf{Black} on the first line is the complete instruction. The several sub-pairs to be annotated are selected by ``$\square$ \#''. The matched pair of the entity phrase and the landmark bounding box are marked with the same color label. The red arrow in the center of the panoramas denotes the next action direction.}
  \label{fig:anno}
\end{figure}

To facilitate the human annotations of entity-landmark grounding, we develop a convenient web-based tool. 
Based on the label-studio platform, we design an annotator-friend interface targeted to our task, as presented in Figure  \ref{fig:anno}. 
\textbf{Black} on the top line is a complete instruction, which consists of several sub-instructions. 
Firstly, the annotators can choose a pair of sub-instruction and sub-trajectory (sub-pair) to be marked sequentially. 
After a sub-pair is selected, the annotators should mark the entity words or phrases in the sub-instruction using different color labels, then mark the matched landmarks in the panoramas using the corresponding color bounding boxes.  
After marking all sub-pairs, the annotators submit the annotations of this episode and mark the next episode. 

\subsection{Annotation Guideline Standardization.}
In the data collection, five individuals with prior experience in visual grounding research served as our experts. 
After building the annotation tool, we adopt the pre-annotation to optimize our tool and standardize the annotation process. 
In the pre-annotation stage, our experts annotate 300 instruction-trajectory pairs together as examples and establish an annotation guideline and several rules based on their consensus after several discussions. 
Four rules are suggested to ensure the standardization of the annotation process.
\begin{compactitem}
     \item \textbf{Alignment Rule}: The entity phrase in the instruction should match the landmark panorama accurately. 
     \item \textbf{Free Text Rule}: Free text instead of the class should be annotated, for instance, ``the white dining table'' instead of ``table''. 
     \item \textbf{Text Coreference Rule}: The entity phrases referring to the same object are marked with the same label.
     \item \textbf{Unique Landmark Rule}: For an entity phrase, only one corresponding landmark bounding box should be annotated in a panorama. 
\end{compactitem}

\subsection{Data Annotation and Revision.}
We first recruit 100 college students to annotate our dataset.
Before starting our task, the students are asked to read the guideline and rules of the annotation carefully and attempt to annotate 50 instruction-trajectory pairs. 
Then we examine each annotated pair if the annotations highly agree with the four rules and reject participators with a low agreement. 
The qualification process leaves us with 43 qualified annotators to complete the annotation task. 
After an annotator finishes the annotations, our experts verify the annotations again and modify the inaccurate part to ensure that the annotations satisfy the four rules. 
In total, the annotation task costs more than 2000 hours and the revision task costs more than 1000 hours. 

\subsection{Data Processing.}
To ensure annotation quality, we first reject the wrong annotations, i.e., alone entity annotations or landmark annotations, and then revise some wrong words in text annotations. 
Due to sub-pairs being annotated,  the obtained positions of entity phrases are based on the sub-instructions. 
So we need to transfer the positions to the corresponding positions in the global instruction. 
On the other hand, we need to transfer the coordinates of the annotated bounding box to the corresponding coordinates in the panorama starting with 0 degrees. 
Finally, we combine the grounded entity-landmark annotations with the R2R dataset, obtaining the Grounded Entity-Landmark R2R (GEL-R2R) dataset. 

\section{Pre-trained Model}
\label{B}

We adopt HAMT \cite{DBLP:conf/nips/ChenGSL21} as our per-trained model, which achieves the SoTA results in many VLN downstream benchmarks. 
Modified from the classical cross-modal model LXMERT \cite{DBLP:conf/emnlp/TanB19}, the HAMT inherits the fully transformer-based architecture.
On the other hand, the HAMT designs a new hierarchical encoder to process history visual observations, which is considered important for decision-making in the long trajectory. Otherwise, to learn more effective initialization for VLN downstream tasks, the model is first pre-trained with several proxy tasks. 

\subsection{Model Architecture.}
The architecture of the pre-trained model is illustrated in Sec. 4.2. 
The pre-trained model takes three inputs: a navigation instruction ${I}$, history information $H_{t}$, and current panoramic visual observation $O_{t}$. 
${I}$ is tokenized by using WordPieces first, and then feed into the language encoder, which is a multi-layer self-attention transformer following the standard BERT, to get a sequence word representation. 
$H_{t}$ consists of all the past panoramic observations $\left\{o_{0,i}, \ldots, o_{t-1, i}\right\}_{i=1}^{36}$ and performed actions $\left\{act_{0}, \ldots, act_{t-1}\right\}$. 
This historic information is input into a history encoder, which has spatial encoding layers and temporal encoding layers. 
The spatiotemporal hierarchical encoder effectively represents history information as $\left\{h_{\mathrm{cls}},h_{0}, \ldots, h_{t-1}\right\}$, where $h_{\mathrm{cls}}$ is to learn a global hidden vector. 
$O_{t}$ consists of image observations $v_{t, i}$ and orientation angle $a_{t,i}$. 
The pre-trained ViT \cite{DBLP:conf/iclr/DosovitskiyB0WZ21} models encode $v_{t, i}$ as a 768-dimensional feature vector, which is concatenated with orientation embedding $\left(\sin \theta_{t, i}, \cos \theta_{t, i}, \sin \phi_{t, i}, \cos \phi_{t, i}\right)$ to obtain the current visual state representation. 
And then cross-modal encoder, composed of self-attention layers and cross-attention layers, jointly encodes the features from the language and vision modality. 
Specifically, the visual modality is the concatenation of history and visual observation.
As a result, the different modalities exchange the signals through cross-attention layers and align the token embedding with the same semantic information. 
Finally, the representations of tokens in instruction, history, and visual state are 
$Z=\left\{z_{\text{cls}}, z_1, \cdots, z_{T}\right\}$, $H_t=\left\{h_{\text{cls}},h_1, \cdots, h_{t-1}\right\}$, $S_t=\left\{s_{1}, \cdots, s_{36}, s_{\text {stop }}\right\}$ respectively. 

\subsection{Pre-training Tasks.}
As studied in previous work, transformer-based models in VLN are commonly pre-trained on the in-domain dataset using several proxy tasks to learn a more effective initialization representation for uni-modal and multi-modal information \cite{DBLP:conf/cvpr/HaoLLCG20, DBLP:journals/corr/abs-2203-11591, DBLP:conf/nips/ChenGSL21}. 
Common vision-language pre-training tasks and the VLN-specific auxiliary tasks are typically served as the proxy tasks. 
The HAMT model is pre-trained by five proxy tasks as follows. 

\paragraph{Masked Language Modeling (MLM) \cite{DBLP:conf/naacl/DevlinCLT19}.}
MLM is a typical pre-training task for BERT-based models. 
In multi-modal transformer-based architecture, the task predicts masked words using surrounding words and image patches. 
It can facilitate the learned word representations to be grounded in the context of visual observations. 
Specifically, with a probability of 15\%, we mask out the input words in the instruction  ${I}$ and replace them with a special token  [MASK]. 
Based on their contextual textual and visual representations, the masked words are predicted via minimizing the negative log-likelihood of original words: 
\begin{equation}
    \mathcal{L}_{\mathrm{MLM}}=-\log p\left(w_m \mid I_{\backslash m}, {H}_T\right),
\end{equation} 
where $I_{\backslash m}$ is the masked instruction, ${H}_T$ is the complete trajectory.

\begin{figure*}
\begin{center}
\setlength{\abovecaptionskip}{0.5cm}
\includegraphics[scale=0.8]{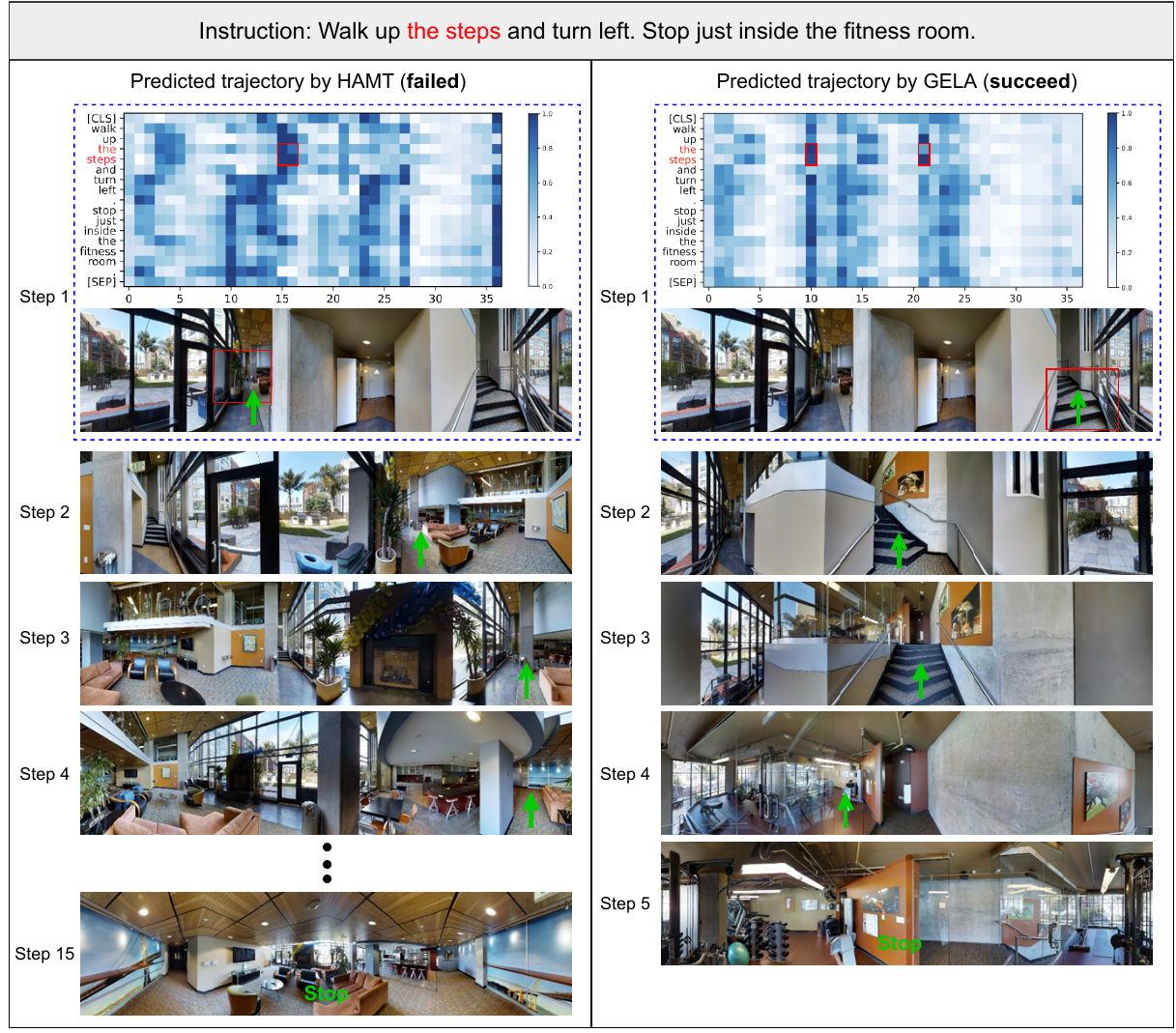}
\caption{Examples in R2R validation unseen split. The green arrow denotes the direction of the next action. Given the instruction on the top line, GELA and HAMT navigate in an environment. In the first step, GELA chooses the true direction but the HAMT chooses the wrong direction. The attention heatmaps at the last transformer layer in the cross-modal encoder are visualized above the panoramas of step 1. In GELA, ``the steps'' attend to the patches of the corresponding landmark (the red bounding box), but ``the steps'' in HAMT attend to other positions in the panorama. Therefore, recognizing ``the steps'' in step 1 helps GELA complete correct navigation.}
\label{fig:ex1}
\end{center}
\end{figure*}

\begin{figure*}\centering
\includegraphics[scale=0.8]{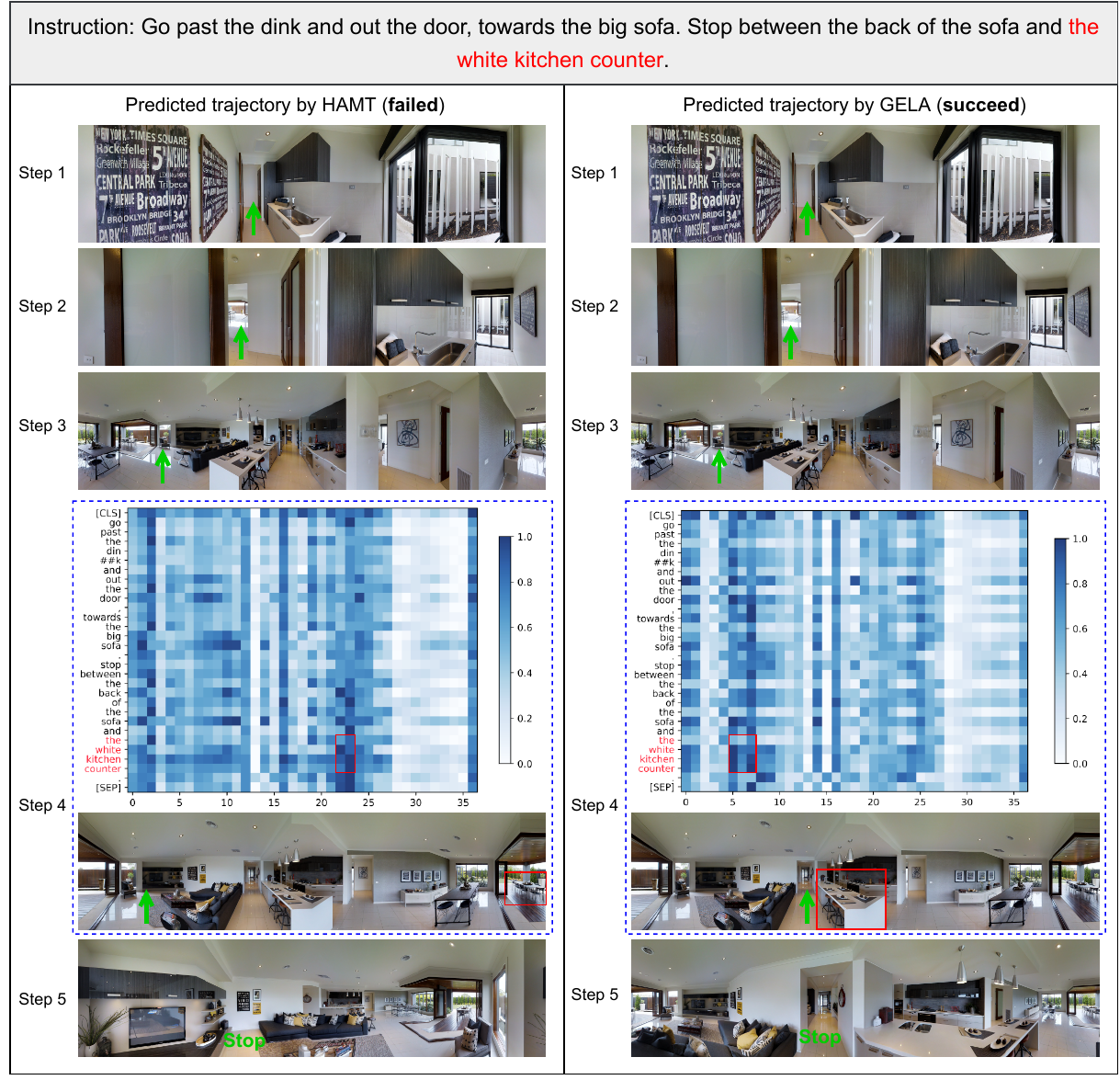}
\caption{Examples in R2R validation unseen split. Given the instruction on the top line, GELA and HAMT navigate in an environment. GELA successfully reaches its destination. In the first three steps, HAMT chooses the right direction. However, in step 4, HAMT makes an error. The attention heatmaps at the last transformer layer in the cross-modal encoder are visualized above the panoramas of step 4. In GELA, ``the white kitchen counter'' attend to the patches of the corresponding landmark (the red bounding box). However, ``the white kitchen counter'' in HAMT attend to another similar landmark in the panorama, which results in the wrong action.}
\label{fig:ex2}
\end{figure*}

\paragraph{Masked Region Classification (MRC) \cite{DBLP:conf/nips/LuBPL19}.}
In analogy to MLM, MRC predicts the semantic class of masked image patches in the panorama based on instruction words and surrounding visual observations. 
It improves the ability of the model to understand the environments and match cross-modal information.
Specifically, we zero out image patches in ${O}_T$ with the probability of 15\% as input. 
For the output embedding of masked patches, we predict the probability distribution $P^{\prime}_i$ on the 1000 classes of ImageNet.  
The objective is to minimize the KL divergence between $P^{\prime}_i$ and the supervisor $P_i$: 
\begin{equation}
    \mathcal{L}_{\text {MRC}}=-\sum_{j=1}^{1000} P_{i, j} \log {P}^{\prime}_{i, j},
\end{equation} 
where $P_i$ is the predicted probability distribution by pre-trained ViT-B/16 \cite{DBLP:conf/iclr/DosovitskiyB0WZ21}.

\paragraph{Instruction Trajectory Matching (ITM) \cite{DBLP:conf/eccv/MajumdarSLAPB20}.}
ITM is a particularly designed task for VLN, which distinguishes whether the input instruction-trajectory pairs match. 
It helps the model to learn the global cross-modal alignment between the instructions and the overall temporal visual trajectory. 
Specifically, we sample four negative trajectories during pre-training for every positive instruction-trajectory pair. 
Two of the negative trajectories are chosen at random from other positive pairs in the mini-batch, and the other two are obtained by temporally rearranging the positive trajectory.
We optimize this task via a Noisy Contrastive Estimation loss \cite{DBLP:journals/jmlr/GutmannH10}: 
\begin{small} 
\begin{equation}
    \mathcal{L}_{\mathrm{ITM}}=-\log \frac{\exp \left(g\left({I}, {H}_T\right)\right)}{\exp \left(g\left({I}, {H}_T\right)\right)+\sum_{k=1}^4 \exp \left(g\left({I}, {H}_{T, k}^{\mathrm{neg}}\right)\right)},
\end{equation} 
\end{small}
where $g\left({I}, {H}_T\right)$ is the global matching score of ${I}$ and ${H}_T$.

\paragraph{Single-step Action Prediction (SAP) \cite{DBLP:conf/nips/ChenGSL21}.}
SAP is a behavior cloning proxy task based on off-line expert demonstrations, which makes the learned representations benefit action decisions. 
The task predicts the next navigation action using instruction, history observations, and the current observation. 
Specifically, we apply a two-layer feedforward network (FFN) to predict action probability for each navigable view: 
\begin{equation}
    p_t\left(s_i^{\prime}\right)=\frac{\exp \left(\operatorname{FFN}\left(s_i^{\prime} \odot z_{\mathrm{cls}}^{\prime}\right)\right)}{\sum_j \exp \left(\operatorname{FFN}\left(s_j^{\prime} \odot z_{\mathrm{cls}}^{\prime}\right)\right)},
\end{equation} 
where $\odot$ is element-wise multiplication and $z_{\mathrm{cls}}$ is the output embedding of the special token [CLS]. 
We optimize this task by minimizing the negative log probability of the target visual state:
\begin{equation}
    \mathcal{L}_{\mathrm{SAP}}=-\log p_t\left(s_{t+1}^{\prime}\right).
\end{equation}

\paragraph{Spatial Relationship Prediction (SPREL) \cite{DBLP:conf/nips/ChenGSL21}.}
SPREL is specially designed for spatial relations in navigation tasks.
The task enhances the competence of the agent to identify directions by learning spatial relation aware representations.
We predict the relative spatial position of two different views in a panorama only based on visual feature $v_i$, angle features $a_i$, or both  $o_i=\left[v_i; a_i\right]$. 
Specifically, we randomly zero out $v_i$ or $a_i$ of the two views with a probability of 30\%. 
The output embeddings of the two views are $o_i^{\prime}$ and $o_j^{\prime}$, and their relative heading and elevation angles are $\theta_{i j}, \phi_{i j}$. 
Then we predict ${\theta}^\prime_{i j}, {\phi}^\prime_{i j}=FFN\left(\left[o_i^{\prime} ; o_j^{\prime}\right]\right)$ .
We optimize this task via minimizing 
\begin{equation}
    \mathcal{L}_{\mathrm{SPREL}}=\left({\theta}^\prime_{i j}-\theta_{i j}\right)^2+\left({\phi}^\prime_{i j}-\phi_{i j}\right)^2.
\end{equation}

\section{Qualitative Examples}
\label{C}

Figure \ref{fig:ex1} and Figure \ref{fig:ex2} show trajectories predicted by our GELA model and compare them to results of the baseline model HAMT \cite{DBLP:conf/nips/ChenGSL21}. 
We see that GELA could better recognize the environment landmarks grounding the corresponding entity phrases in instructions.

\end{document}